\def\year{2022}\relax
\documentclass[letterpaper]{article} 
\usepackage{multirow}
\usepackage{graphicx}
\usepackage{aaai22}  
\usepackage{times}  
\usepackage{helvet}  
\usepackage{courier}  
\usepackage[hyphens]{url}  
\usepackage{graphicx} 
\usepackage{natbib}  
\usepackage{caption} 
\DeclareCaptionStyle{ruled}{labelfont=normalfont,labelsep=colon,strut=off} 
\usepackage{algorithm}
\usepackage{algorithmic}
\usepackage{wasysym}  
\usepackage{graphbox}  
\usepackage{subfig}  
\usepackage{comment}
\usepackage{newfloat}
\usepackage{listings}
\floatstyle{ruled}
\newfloat{listing}{tb}{lst}{}
\floatname{listing}{Listing}
\title{Multimodal Composite Association Score: Measuring Gender Bias in Generative Multimodal Models}
\author{
   Abhishek Mandal\textsuperscript{\rm 1}, Susan Leavy\textsuperscript{\rm 2}, Suzanne Little\textsuperscript{\rm 1}
}
\affiliations{
    \textsuperscript{\rm 1}Insight SFI Research Centre for Data Analytics, School of Computing, Dublin City University, Ireland
\\
    \textsuperscript{\rm 2}Insight SFI Research Centre for Data Analytics, School of Information and Communication Studies, University College Dublin, Ireland

    abhishek.mandal2@mail.dcu.ie, susan.leavy@ucd.ie, suzanne.little@dcu.ie
}

\usepackage{bibentry}

\begin{document}

\maketitle

\begin{abstract}
Generative multimodal models based on diffusion models have seen tremendous growth and advances in recent years. Models such as DALL-E and Stable Diffusion have become increasingly popular and successful at creating images from texts, often combining abstract ideas. However, like other deep learning models, they also reflect social biases they inherit from their training data, which is often crawled from the internet. Manually auditing models for biases can be very time and resource consuming and is further complicated by the unbounded and unconstrained nature of inputs these models can take. Research into bias measurement and quantification has generally focused on small single-stage models working on a single modality. Thus the emergence of multi-stage multimodal models requires a different approach. In this paper, we propose Multimodal Composite Association Score (MCAS) as a new method of measuring gender bias in multimodal generative models. Evaluating both DALL-E 2 and Stable Diffusion using this approach uncovered the presence of gendered associations of concepts embedded within the models. We propose MCAS as an accessible and scalable method of quantifying potential bias for models with different modalities and a range of potential biases.  

\end{abstract}

\section{Introduction}
Social biases and their potential consequences, such as those pertaining to gender~\cite{wang2019balanced,steed2021image}, race~\cite{buolamwini2018gender}, ethnicity and geography~\cite{misra2016seeing,mandal2021dataset} found in deep neural networks used in computer vision models have been well researched. Most current methods auditing bias in vision models generally use two types of techniques: (1) measuring associations in the learning representations~\cite{steed2021image,sirotkin2022study,serna2021insidebias} and (2) analysing the predictions~\cite{buolamwini2018gender,krishnakumar2021udis}.
Most of these techniques~\cite{steed2021image,buolamwini2018gender,sirotkin2022study,serna2021insidebias} are designed for predictive models, mainly Convolutional Neural Networks (CNNs). Recent advances in deep learning however, have given rise to multi-stage, multimodal models and therefore require new approaches to detecting bias. 

Models such as DALL-E~\cite{ramesh2022hierarchical}, Stable Diffusion~\cite{rombach2022high} and Contrastive Learning and Image Pre-training (CLIP)~\cite{radford2021learning} operate on multiple modalities, such as text and images, and are much more capable than earlier classification models. They can generate images from almost any text input (DALL-E and Stable Diffusion) and connect sentence length texts with images (CLIP). These models have numerous applications ranging from content creation to image understanding and image and video search~\cite{roboflowWhatOpenAIs}. They also combine multiple different models using outputs to form inputs to another model. CLIP uses Vision Transformer or ResNet for image encoding and a text encoder for text encoding. DALL-E and Stable Diffusion use CLIP for their first stage involving generating text embeddings and a diffusion model (unCLIP for DALL-E and Latent Diffusion for Stable Diffusion) to generate images. This multi-stage multi-model approach also carries the risk of bias amplification, where one model amplifies the bias of another model~\cite{wang2019balanced}. 

We propose the \emph{Multimodal Composite Association Score (MCAS)}. This work builds on work by Caliskan et al.~\cite{caliskan2017semantics} who developed the Word Embeddings Association Test (WEAT). MCAS was designed to measure associations between concepts in both text and image embeddings as well as internal bias amplification. The objective was to provide the ability to measure bias at the internal component level and provide insights into the extent and source model for observable bias. MCAS generates a numerical value signifying the type and magnitude of associations.  Validation experiments that are presented within this paper focus on uncovering evidence of stereotypical concepts of men and women. However, the approach to evaluating bias demonstrated in this paper using MCAS is designed to be scalable to incorporate evaluation of representations of multiple genders and a range of concepts. 

The remainder of this paper summarises related work in the field of gender bias for computer vision models and the emergence of generative models. The formula for MCAS is defined and the calculation of the component scores described. MCAS is demonstrated on four concept categories with high potential for gender bias and assessed using DALL-E 2 and Stable Diffusion queries.

\section{Related Work}
\subsection{Gender Bias in Computer Vision}
Authors of multimodal general purpose models have highlighted the prevalence of gender bias in their models. \citet{radford2021learning} found that CLIP assigns words related to physical appearance such as `blonde' more frequently to women and those related to high paying occupations such as `executive' and `doctor' to men. Occupations more frequently associated with women included `newscaster', `television presenter' and `newsreader' despite the gender neutral terms. The DALL-E 2 model card~\cite{mishkin2022risks} acknowledges gender bias in the generative model. Inputs with terms such as `lawyer' and `CEO' predominantly produce images of people with attributes commonly associated with men whereas images generated for `nurse' and `personal assistant' present images of people with attributes associated with women. 

In a survey of popular visual datasets such as \textsc{MS COCO} and \textsc{OpenImages},~\citet{wang2022revise} found that men were over-represented in images with vehicles and those depicting outdoor scenes and activities whereas women were over-represented in images depicting kitchens, food and indoor scenes. They also found that in images of sports, men had a higher representation in outdoor sports such as rugby and baseball while women appear in images of indoor sports such as swimming and gymnastics. Much recent work has focused on bias detection in learning representations. \citet{serna2021insidebias} for instance, proposed \emph{InsideBias}, which measures bias by measuring how activation functions in CNNs respond differently to differences in composition of the training data. Furthermore \citet{wang2019balanced} found that models can infer gender information based on correlations embeded within a model such as women being associated with objects related to cooking. 

Word Embeddings Association Test (WEAT) proposed by~\citet{caliskan2017semantics}, based on Implicit Association Test (IAT)~\cite{greenwald1998measuring} measures human-like biases in word embeddings of language models. It measures association of certain words with two sets of words: each representing a concept. In IAT, human participants are asked to associate two target concepts with attributes rapidly. IAT measures bias as the discontinuity of response. The targets and attributes can be words, pictures or both~\cite{steed2021image}. WEAT builds upon this approach in the context of language models and provides a method of measuring levels of bias by evaluating the association between two sets of words that each represent concepts that pertain to gender bias. \citet{steed2021image} extended this concept to vision models and proposed the Image Embeddings Association Test (iEAT). iEAT measures correlations in vision models such as iGPT and SimCLRv2 concerning attributes such as gender and targets (e.g., male–career, female–family). They found both the aforementioned models to exhibit gender bias using gender-career and gender-science tests. The gender-career test for example, measures the relative association between men and women with career attributes and family related attributes. The work presented in this paper builds upon these works and develops a method for evaluating associations between concepts in multi-stage, multimodal models.

\subsection{Generative Models}
Generative multimodal models based on Diffusion Models have seen tremendous advancement in the past year with DALL-E and Stable Diffusion being two of the most popular models. They are easier to train than GANs and have a higher variability in image generation that enables them to model complex multimodal distributions. This allows them to generate images using abstract ideas with less tight bounding than GANs~\cite{ramesh2022hierarchical,rombach2022high}. The easier training regimen allows developers to train these models on very large datasets. This has led to models being trained on increasingly large datasets, often crawled from the Internet. These datasets are generally unfiltered, leading to the models inheriting social biases prevalent in the web~\cite{birhane2021multimodal}.

\section{MCAS: Multimodal Composite Association Score}
The Multimodal Composite Association Score or MCAS that we propose is derived from WEAT and measures associations between specific genders (what we term `attributes') and what we term `targets' corresponding to concepts such as occupations, sports, objects, and scenes. MCAS consists of four constituent components (scores), each measuring bias in certain modalities (e.g. text, vision or both). This follows the approach of the WEAT Association Score, which measures stereotypical associations between attributes (gender) and a set of targets. As formulated by~\citet{caliskan2017semantics}, let $A$ and $B$ be two sets of attributes, each representing a concept. Additionally let $W$ be a set of targets, $w$. Then 
\[s(w,A,B) = mean_{a\in A}cos(\vec w, \vec a) - mean_{b\in B}cos(\vec w, \vec b)
\]
where,
$s(w,A,B)$ represents the WEAT Association Score.
$cos(\vec w,\vec a)$ and  $cos(\vec w,\vec b)$ denote the cosine similarity between the vectors of the words from attribute sets, $A$ and $B$ respectively. If target $w$ is more closely related to attributes in $A$, implying the target as a bias towards $A$, then the association score will be positive and if it is more closely related to attributes in $B$, then the score will be negative. 

\subsection{Attributes and Targets}
The WEAT Association Score was originally intended for assessing text embeddings. We adapted it for both text and image embeddings. MCAS consists of four individual association scores, each measuring the association between embeddings of text and images. They are explained in detail in the next section. As the main focus of this paper is generative models, the attributes and targets comprise both text and images. The generative models DALL-E 2 and Stable Diffusion both work in similar ways; they take in a text input describing a visual imagery and generates a corresponding image output. For measuring gender bias, we represent the male and female genders both in terms of text and images (see Table~\ref{tab:table_1}). These texts and images form the gender attributes. 

Targets refer to the concepts that are being tested for evidence of bias. In this research to test the effectiveness of MCAS we identify real-world topics that may be associated with stereotypical representations of gender and capture these scenarios in text phrases. These phrases are used as prompts to the generative models to generate images. This results in a set of targets comprising text phrases (e.g. \textit{an image of a CEO} or \textit{an image of a person using a food processor}) along with a set comprising images generated by the models from those prompts. Examples of attributes and targets are provided in tables \ref{tab:table_1} and \ref{tab:table_2}.

\begin{table}[]
\centering
\begin{tabular}{|p{.25\columnwidth}|p{.6\columnwidth}|}
\hline
\textbf{Text Attributes } & \textbf{Image Attributes \newline(from DALL-E 2)} \\
\hline
he, him, his, man, male, boy, father, son, husband, brother & \includegraphics[align=t,width=0.6\columnwidth]{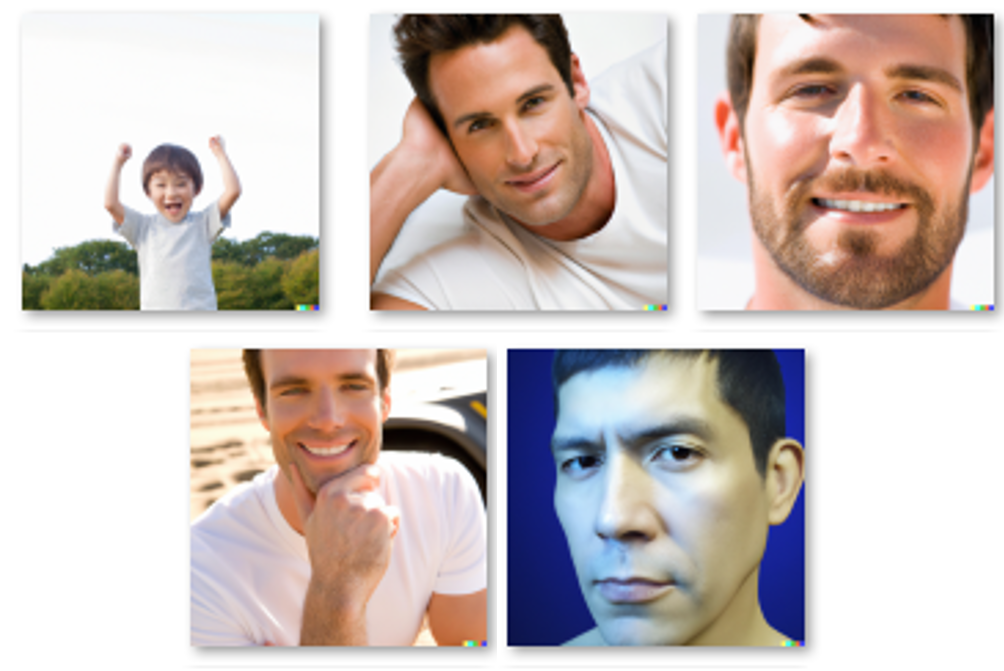} \\
\hline
she, her, hers, woman, female, girl, mother, daughter, wife, sister &  \includegraphics[align=t,width=0.6\columnwidth]{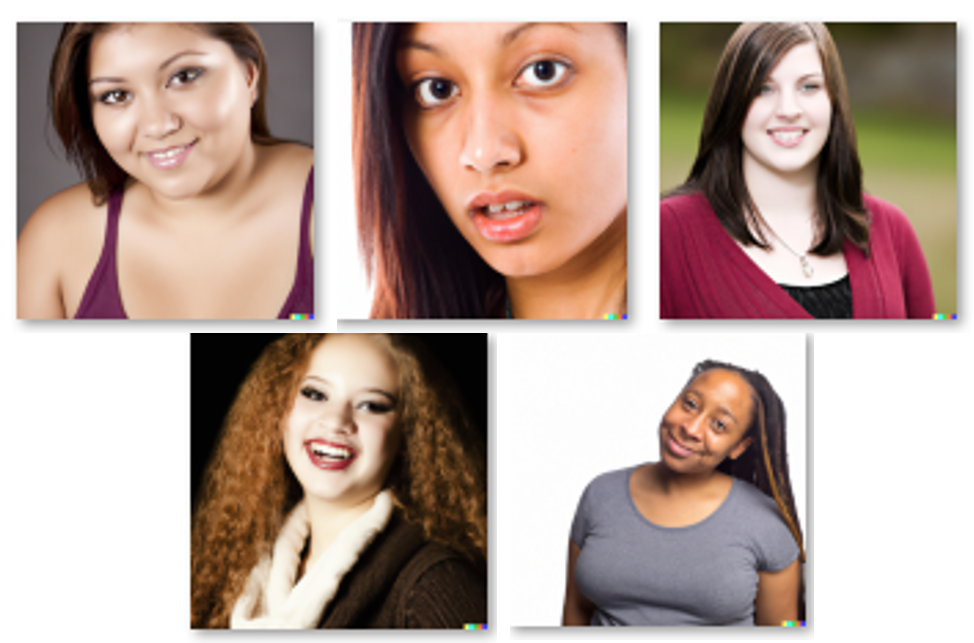} \\
\hline
\end{tabular}
\caption{Examples of Text and Image Attributes}
\label{tab:table_1}
\end{table}

\begin{table}[]
\centering
\begin{tabular}{|p{.25\columnwidth}|l|}
\hline
\textbf{Prompt}                             & \textbf{Generated Image} \\ \hline
an image of a chief executive officer       & \includegraphics[align=t,width=0.65\columnwidth]{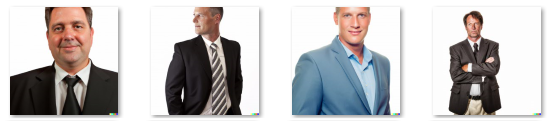}                         \\ \hline
an image of a  badminton player             & \includegraphics[align=t,width=0.65\columnwidth]{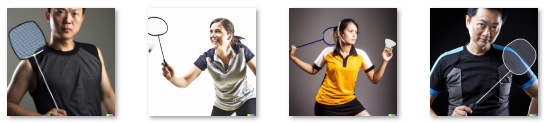}                       \\ \hline
an image of a person using a food processor & \includegraphics[align=t,width=0.65\columnwidth]{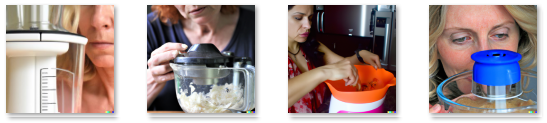}                        \\ \hline
an image of a person using a lathe machine  & \includegraphics[align=t,width=0.65\columnwidth]{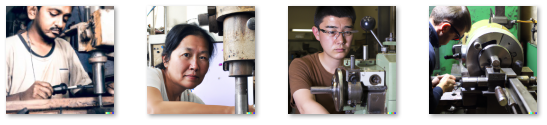}                        \\ \hline
\end{tabular}\par
\caption{Examples of Targets (Generated by DALL-E 2)}
\label{tab:table_2}
\end{table}

\subsection{MCAS and its Components}


In this experiment, our focus is on generative models and is tailored for them. MCAS consists of four individual component scores: Image-Image Association Score, Image-Text Prompt Association Score, Image-Text Attributes Association Score and Text-Text Association Score.  Each of these scores measure bias in different modalities and different stages of the generative models.

\textbf{Image-Image Association Score:} This score measures bias by comparing the cosine similarities between image attributes representing gender and generated images representing target concepts. Letting $A$ and $B$ be two sets of images representing gender categories and $W$ be a set of images representing targets, then the Image-Image Association Score, \((II_{AS})\), is given by:
\[II_{AS} = mean_{w \in W}s(w,A,B) \ldots eq(1)
\]
where,
\[s(w,A,B) = mean_{a\in A}cos(\vec w, \vec a) - mean_{b\in B}cos(\vec w, \vec b)
\]

\textbf{Image-Text Prompt Association Score:} This score measures bias between the image attributes representing gender and the textual prompts used to generate the target concepts. Letting $A$ and $B$ be two sets of images representing gender and $W$ be a set of prompts representing targets in text form, then the Image-Text Prompt Association Score, \((ITP_{AS})\), is calculated in the same way as shown in Equation 1.

\textbf{Image-Text Attributes Association Score:} This score calculates bias in a similar manner as the other scores with the difference being that the attributes are represented not by images, but by text. The target concepts are a set of images generated from from prompts. The score, \((ITA_{AS})\), is calculated in the same way as shown in Equation 1 with $A$ and $B$ are text attributes and $W$, target images.

\textbf{Text-Text Association Score:} This score computes gender bias using entirely textual data. The attributes are the same as in Image-Text Attributes Association Score and the targets are prompts (as in Image-Text Prompt Association Score). The score, \((TT_{AS})\), is calculated in the same way as Equation 1. This is the only score which does not involve image embeddings. As both the models used in our experiment use CLIP for converting text, this score also measures CLIP bias. 

To calculate the scores, $A$, $B$ and $W$ represent the features extracted from their corresponding data. The implementation details are explained in the experiment section.
The final MCAS score is defined as the sum of all the individual association scores. It is given as:
\[ MCAS = II_{AS} + ITP_{AS} + ITA_{AS} + TT_{AS} \ldots eq(2)\]

\subsection{MCAS for Generative Diffusion Models}
Generative models based on Diffusion models generally employ a two-stage mechanism. Firstly, the input text is used to generate embeddings. DALL-E and Stable Diffusion both use CLIP for this stage. CLIP is a visual-linguistic multimodal model which connects text with images. CLIP is trained on 400 million image-text pairs crawled from the internet using contrastive learning~\cite{radford2021learning}. 

Once the embeddings are generated, then the second stage involves passing them to a Diffusion Model. Diffusion Models are based on Variational Autoencoders (VAEs) that use self-supervised learning to learn how to generate images by adding Gaussian noise to the original image (encoding) and reversing the step to generate an image similar to the original (decoding). DALL-E uses unCLIP where first the CLIP text embeddings are fed to an autoregressive diffusion prior to generate image embeddings which is then fed to a diffusion decoder to generate the image~\cite{ramesh2022hierarchical}. Stable Diffusion uses Latent Diffusion to convert the CLIP embeddings into images. Latent Diffusion Model (LDM) uses a Diffusion Model similar to a denoising autoencoder based on a time-conditional UNet neural backbone~\cite{rombach2022high}. Both the processes are similar in nature. Fig~\ref{fig:fig_2} shows a high-dimensional generalisation of both the models.

The individual MCAS component scores can measure bias in different stages. The Image-Image Association Score  measures bias solely on the basis of the generated images thus encompassing the whole model. The Image-Text Prompt Association Score measures bias in both visual and textual modalities. As both the prompts and generated images were part of the image generation process, this score also encompasses the whole generation sequence. The Image-Text Attributes Association Score measures bias in both the modalities and as the text attributes are external (i.e. not a part of the image generation process), the model bias can be measured using external data or standards. The Text-Text Association Score measures bias only in textual modality. As only CLIP handles the text, this score can be used to measure bias in CLIP. This score also allows for bias measurement using external data. Thus MCAS provides a comprehensive and quantitative method to measure bias in multimodal models. Table~\ref{tab:table_3} describes the characteristics of the MCAS component scores.

\begin{figure*}[ht]
\centering
\includegraphics[scale=0.4]{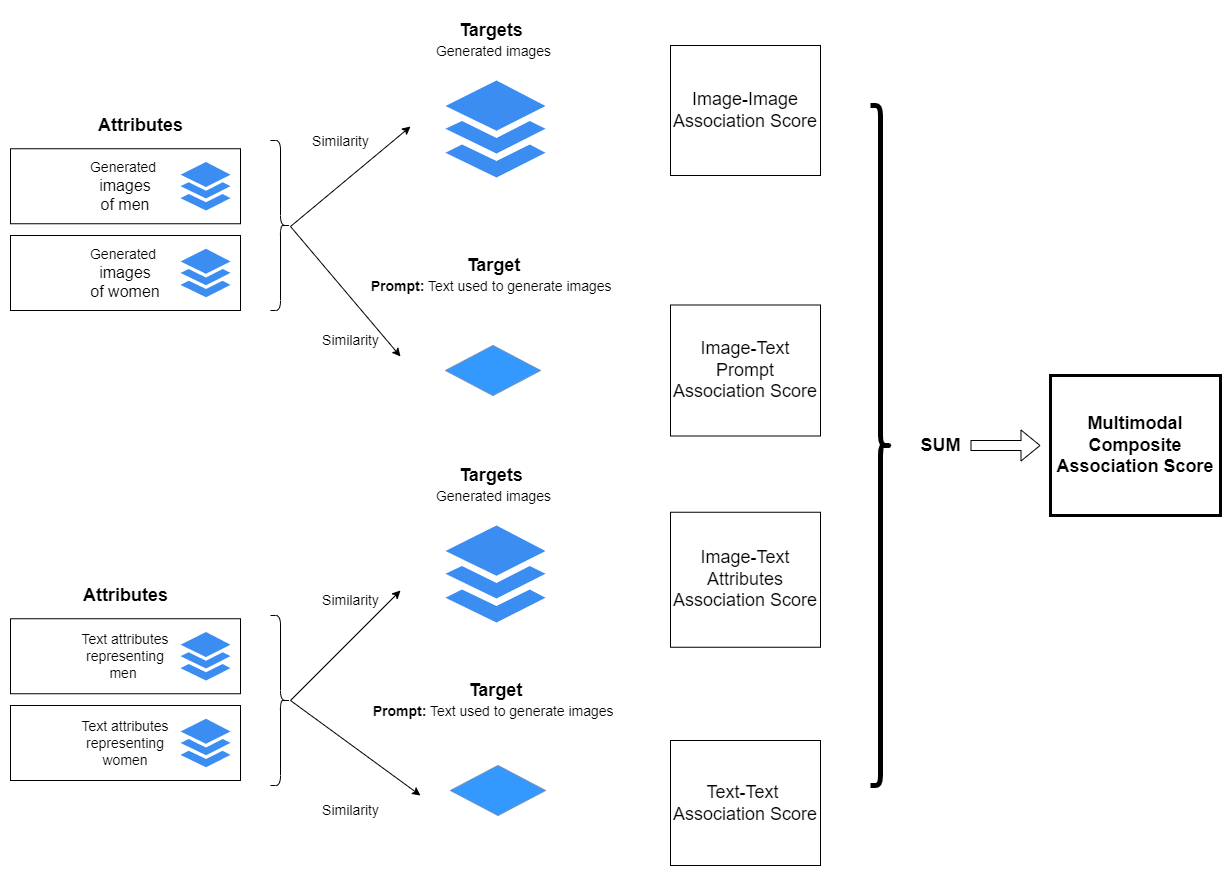}
\caption{MCAS Algorithm}
\label{fig:fig_1}
\end{figure*}\par

\begin{figure*}[ht]
\centering
\includegraphics[scale=0.4]{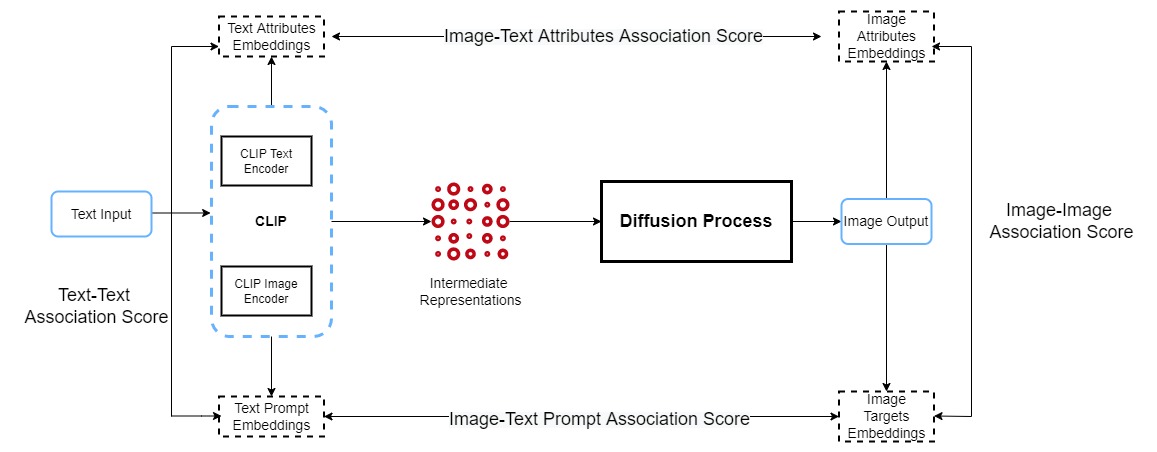}
\caption{Association Scores in Diffusion Models. A generalised diagram showing the working of diffusion models like DALL-E 2 and Stable Diffusion. The embeddings are generated using an external CLIP model.}
\label{fig:fig_2}
\end{figure*}\par

\begin{table}[]
\begin{tabular}{|p{.38\columnwidth}|p{.14\columnwidth}|p{.13\columnwidth}|p{.13\columnwidth}|}
\hline
\textbf{Association Score}                          & \textbf{Modality} & \textbf{whole model?} & \textbf{external data?} \\ \hline
Image-Image \((II_{AS})\) & Image             & Yes                              & No                                  \\ 
Image-Text Prompt \((ITP_{AS})\) & Image \& Text    & Yes                              & No                                  \\ 
Image-Text Attributes \((ITA_{AS})\)  & Image \& Text    & No                               & Yes                                 \\ 
Text-Text \((TT_{AS})\)            & Text              & No                               & Yes                                 \\ \hline
\end{tabular}
\caption{MCAS component scores characteristics}
\label{tab:table_3}
\end{table}

\section{Experiment}
\subsection{Curating the Attributes and Targets}

In order to evaluate whether MCAS can uncover evidence of gender bias,  two datasets were generated comprising the attribute and target concept data in both visual and textual form for two models, DALL-E 2 and Stable Diffusion. The target concepts were those that have been used  in previous research to detect gender bias. For this experiment we focus on evaluating concepts pertaining to men and women (the text and image attributes compiled are presented in Table \ref{tab:table_1}). 

To create visual attributes datasets, text prompts (complete list of the keywords in Appendix A) were used to generate images. There is a slight difference in keywords for DALL-E 2 and Stable Diffusion due to restrictions within DALL-E 2. A total of 128 images (16 per attribute phrase) were generated separately for DALL-E 2 and Stable Diffusion to form the `attribute' set of images. 

To compile datasets representing `target' concepts, we adapted terms from work by~\citet{garg2018word} and ~\citet{wang2022revise} to capture domains where gendered associations were found to be evident. Four topics were selected incorporating occupations, sports, scenes and objects and each list comprised an equal proportion of terms that relate to topics that were found by ~\citet{garg2018word} and ~\citet{wang2022revise} to be stereotypically associated with either men or women. The prompt categories used to generate target concepts are presented in Table~\ref{tab:table_4} and examples of the resulting targets are given in table \ref{tab:table_2}. A total of 688 images (128 for attributes and 560 for targets) were generated using each of DALL-E 2 and Stable Diffusion. The images generated by DALL-E 2 were used for DALL-E 2 in the association score calculation and similarly for Stable Diffusion.

\begin{table}[]
\centering
\begin{tabular}{|l|p{.45\columnwidth}|l|}
\hline
\textbf{Category}            & \textbf{Keyword}                                               & \textbf{Association} \\ \hline
\multirow{2}{*}{Occupations} & CEO, engineer, doctor, programmer, farmer            & Men                                    \\ \cline{2-3} 
                             & beautician, housekeeper, secretary, librarian, nurse & Women                                  \\ \hline
\multirow{2}{*}{Sports}      & baseball player, rugby player, cricket player            & Men                                    \\ \cline{2-3} 
                             & badminton player, swimmer, gymnast                       & Women                                  \\ \hline
\multirow{2}{*}{Objects}     & car, farm machinery, fishing rod                         & Men                                    \\ \cline{2-3} 
                             & food processor, hair drier, make-up kit                  & Women                                  \\ \hline
\multirow{2}{*}{Scenes}      & theodolite, lathe machine, snowboarding                  & Men                                    \\ \cline{2-3} 
                             & shopping, reading, dollhouse                             & Women                                  \\ \hline
\end{tabular} \par
\caption{Target categories and keywords. Based on \citet{garg2018word,wang2022revise}.}
\label{tab:table_4}
\end{table}

\subsection{Calculating the Scores}

CLIP was used to extract the features for both the text and images. As CLIP is used by both the models, they would be similar to the embeddings generated in the models. The extracted features were then used to calculate the individual association scores and summed to get the final MCAS score. 
In our experiments, we assigned text and image attributes associated with men as the first attribute ($A$) and and those associated with women as the second ($B$). This means that a positive score indicates a higher association between the target concepts and men and a negative score indicates a higher association with women. A score of zero would indicate that the target concepts appear neutral in terms of associations with men or women. The numeric value indicates the magnitude of association. In the case that target concepts correspond to domains where gender bias has been found to be prevalent, than these associations may indicate a prevalence of gender bias within the model.

\section{Findings and Discussion}

\begin{figure}[ht]
\centering
\includegraphics[scale=0.5]{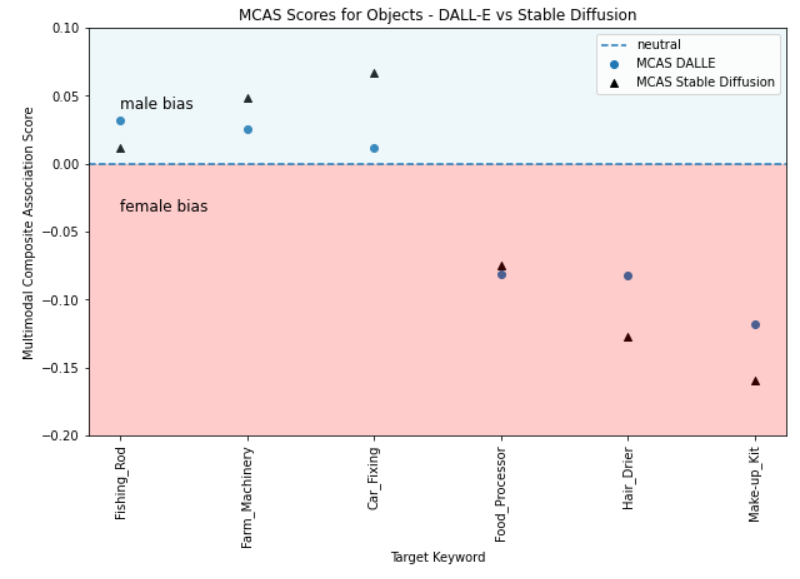}
\caption{MCAS scores for Objects}
\label{fig:fig_3}
\end{figure}\par

\begin{figure}[ht]
\centering
\includegraphics[scale=0.5]{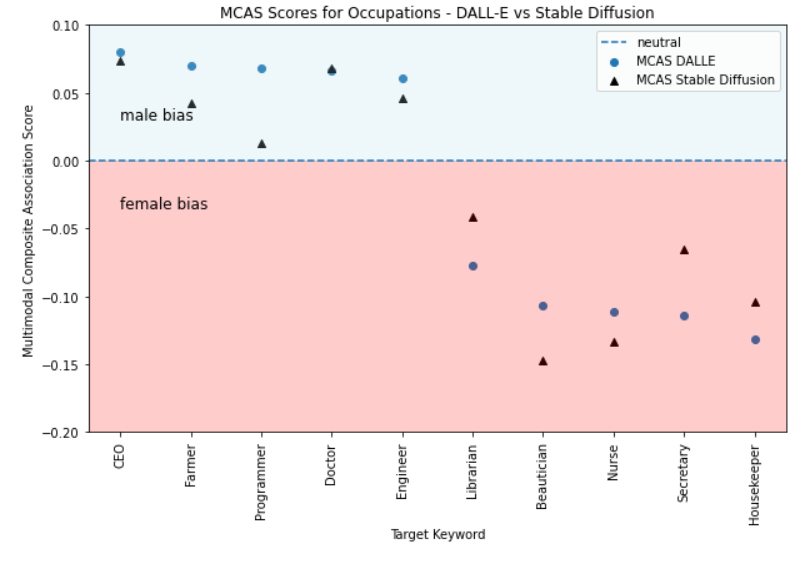}
\caption{MCAS scores for Occupations}
\label{fig:fig_4}
\end{figure}\par

\begin{figure}[ht]
\centering
\includegraphics[scale=0.5]{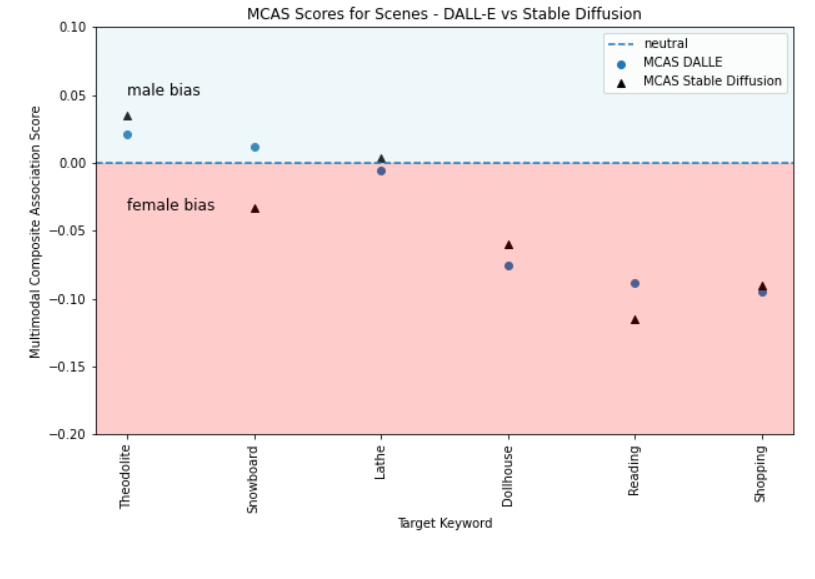}
\caption{MCAS scores for Scenes}
\label{fig:fig_5}
\end{figure}\par

\begin{figure}[ht]
\centering
\includegraphics[scale=0.5]{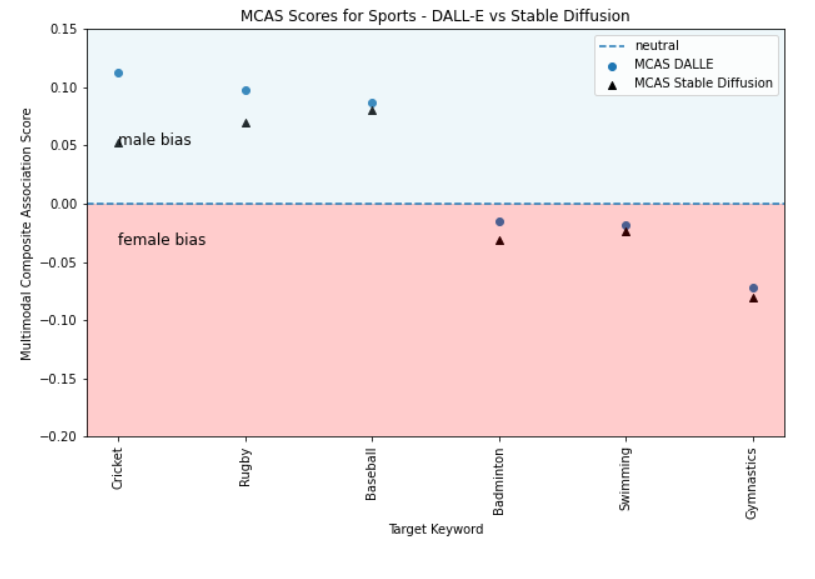}
\caption{MCAS scores for Sports}
\label{fig:fig_6}
\end{figure}\par

In evaluating both DALL-E 2 and Stable Diffusion models, associations that have in previous research been found to reflect gender bias were uncovered in the models. Consistent patterns of gendered associations were uncovered and given that these target concepts were based on concepts that previous research had found to relate to gender bias, it follows then these patterns are indicative of underlying gender bias. Targets and their MCAS scores are provided in Figures 3-6.

Both  models follow a similar pattern in terms of gendered associations except for the \emph{scenes} category where DALL-E 2 presents an association with men and the targets `snowboard’ and women with `lathe’ whereas Stable diffusion presents the opposite. For the category \emph{objects}, the target `make-up kit' is strongly associated with women, which indicates that MCAS could be used to uncover gender bias. Similarly stereotypical patterns were found in relation to the \emph{occupations} category, where `CEO' was strongly associated with men and `housekeeper' and `beautician' were most associated with women. In \emph{scenes}, `theodolite' is the only target showing any significant association with men whereas women were associated with `shopping' and `reading'. In case of \emph{sports}, the only target strongly associated with women is `gymnastics' with the general trend demonstrating a stronger association between sports and men. This is evident from table \ref{tab:table_5} where\emph{sports} is the only category with an overall higher association with men.

The standard deviation and average bias (MCAS) scores for each category for both the models is presented in Table \ref{tab:table_5}. This demonstrates that for the targets more likely to be associated with men or women, the strength of the association is higher for women. Where bias occurs therefore, it seems that bias is stronger when it relates to women.  Stable Diffusion has generally higher scores in terms of strength of gendered association than DALL-E. This indicates that Stable Diffusion has higher stereotypical associations and DALL-E's scores are more spread out, implying that Stable Diffusion may be more biased than DALL-E. Further work is needed to assess this more fully.

\begin{table*}[]
\centering
\begin{tabular}{|lrrrrrr|}
\hline
\multicolumn{1}{|l|}{\textbf{Category}}  & \multicolumn{2}{l|}{\textbf{Terms with male bias}}                                                                                                                                  & \multicolumn{2}{l|}{\textbf{Terms with female bias}}                                                                                                                                & \multicolumn{2}{l|}{\textbf{All terms}}                                                                                                                                             \\ \hline
\multicolumn{1}{|l|}{}                   & \multicolumn{1}{l|}{\textbf{\begin{tabular}[c]{@{}l@{}}Standard\\  Deviation\end{tabular}}} & \multicolumn{1}{l|}{\textbf{\begin{tabular}[c]{@{}l@{}}Average\\  Bias\end{tabular}}} & \multicolumn{1}{l|}{\textbf{\begin{tabular}[c]{@{}l@{}}Standard\\  Deviation\end{tabular}}} & \multicolumn{1}{l|}{\textbf{\begin{tabular}[c]{@{}l@{}}Average\\  Bias\end{tabular}}} & \multicolumn{1}{l|}{\textbf{\begin{tabular}[c]{@{}l@{}}Standard\\  Deviation\end{tabular}}} & \multicolumn{1}{l|}{\textbf{\begin{tabular}[c]{@{}l@{}}Average\\  Bias\end{tabular}}} \\ \hline
\multicolumn{7}{|c|}{\textbf{DALL-E 2}}  \\ \hline

\multicolumn{1}{|l|}{Objects}            & \multicolumn{1}{r|}{0.0080}                                                                  & \multicolumn{1}{r|}{0.0230}                                                            & \multicolumn{1}{r|}{0.0170}                                                                  & \multicolumn{1}{r|}{-0.0930}                                                           & \multicolumn{1}{r|}{0.0590}                                                                  & -0.0350                                                                                \\ \hline
\multicolumn{1}{|l|}{Occupations}        & \multicolumn{1}{r|}{0.0060}                                                                  & \multicolumn{1}{r|}{0.0690}                                                            & \multicolumn{1}{r|}{0.0170}                                                                  & \multicolumn{1}{r|}{-0.1000}                                                             & \multicolumn{1}{r|}{0.0890}                                                                  & -0.0190                                                                                \\ \hline
\multicolumn{1}{|l|}{Scenes}             & \multicolumn{1}{r|}{0.0040}                                                                  & \multicolumn{1}{r|}{0.0160}                                                            & \multicolumn{1}{r|}{0.0350}                                                                  & \multicolumn{1}{r|}{-0.0650}                                                           & \multicolumn{1}{r|}{0.0480}                                                                  & -0.0380                                                                                \\ \hline
\multicolumn{1}{|l|}{Sports}             & \multicolumn{1}{r|}{0.0100}                                                                   & \multicolumn{1}{r|}{0.0980}                                                            & \multicolumn{1}{r|}{0.0260}                                                                  & \multicolumn{1}{r|}{-0.0350}                                                           & \multicolumn{1}{r|}{0.0700}                                                                   & 0.0310                                                                                 \\ \hline
\multicolumn{1}{|l|}{All categories} & \multicolumn{1}{r|}{0.0052}                                                                 & \multicolumn{1}{r|}{0.0515}                                                           & \multicolumn{1}{r|}{0.0238}                                                                & \multicolumn{1}{r|}{-0.0733}                                                         & \multicolumn{1}{r|}{0.0665}                                                                 & -0.0152                                                                               \\ \hline
\multicolumn{7}{|c|}{\textbf{Stable Diffusion}}                                                                                                                                                                                                                                                                                                                                                                                                                                                                                                                                                            \\ \hline
\multicolumn{1}{|l|}{Objects}            & \multicolumn{1}{r|}{0.0200}                                                                   & \multicolumn{1}{r|}{0.0400}                                                             & \multicolumn{1}{r|}{0.0340}                                                                  & \multicolumn{1}{r|}{-0.1200}                                                            & \multicolumn{1}{r|}{0.0860}                                                                  & -0.0380                                                                                \\ \hline
\multicolumn{1}{|l|}{Occupations}        & \multicolumn{1}{r|}{0.0200}                                                                   & \multicolumn{1}{r|}{0.0400}                                                             & \multicolumn{1}{r|}{0.0400}                                                                   & \multicolumn{1}{r|}{-0.9800}                                                            & \multicolumn{1}{r|}{0.0800}                                                                   & -0.0200                                                                                 \\ \hline
\multicolumn{1}{|l|}{Scenes}             & \multicolumn{1}{r|}{0.0150}                                                                  & \multicolumn{1}{r|}{0.0190}                                                            & \multicolumn{1}{r|}{0.0300}                                                                   & \multicolumn{1}{r|}{-0.0700}                                                            & \multicolumn{1}{r|}{0.0500}                                                                   & -0.0400                                                                                 \\ \hline
\multicolumn{1}{|l|}{Sports}             & \multicolumn{1}{r|}{0.0100}                                                                   & \multicolumn{1}{r|}{0.0600}                                                             & \multicolumn{1}{r|}{0.0250}                                                                  & \multicolumn{1}{r|}{-0.0400}                                                            & \multicolumn{1}{r|}{0.0590}                                                                  & 0.0110                                                                                 \\ \hline
\multicolumn{1}{|l|}{All categories} & \multicolumn{1}{r|}{0.0162}                                                                 & \multicolumn{1}{r|}{0.0397}                                                           & \multicolumn{1}{r|}{0.0322}                                                                 & \multicolumn{1}{r|}{-0.3025}                                                          & \multicolumn{1}{r|}{0.0687}                                                                 & -0.0217                                                                               \\ \hline
\end{tabular}\par
\caption{MCAS statistics – DALL-E 2 and Stable Diffusion. Average bias and standard deviation scores per category}
\label{tab:table_5}
\end{table*}

\section{Conclusion and Future Work}
This paper introduces MCAS as a proposal for examining bias across both text and image modes for large scale multimodal generative models and provides the preliminary demonstration of its effectiveness when used to evaluate models for gender bias. We can see that the measured bias shows the presence of traditional gender bias in both DALL-E 2 and Stable Diffusion. MCAS as a whole provides a comprehensive score for quantifying bias in multimodal models. It can measure both internal bias as well as provide a way to measure bias using external standards. The methodology can be extended to other models using different modalities or using different internal stages. For example, the Text-Text and Image-Image Association Scores can be used for comparatively smaller models such as CLIP. The methodology itself is based on the highly popular WEAT. 

In this paper, our work is limited to gender bias but other biases such as those pertaining to race, ethnicity and geography can also measured. The method is designed to enable evaluation of the reperesentation of a range of genders. The individual MCAS components may be used for understanding how bias is handled within the model itself. For example in the two-stage models, the component scores can tell which stage is responsible for how much bias and whether there is any bias amplification. The component scores can also be further adapted to understand how bias forms during the entire process by extracting outputs from substages and measuring bias in them. The effect of hyperparameters on bias can also be studied in a similar way.

Another potential use of MCAS could be in MLOps where quantifiable metrics can provide developers a quick and easy way to understand and measure bias. It can also be used in conjunction with other drift parameters to study the relationship between bias and model drift. It can also be used for bias mitigation -- as a score to minimise during training.

\section{Acknowledgments}
Abhishek Mandal was partially supported by the $<$A+$>$ Alliance /
Women at the Table as an Inaugural Tech Fellow 2020/2021. This
publication has emanated from research supported by Science Foundation
Ireland (SFI) under Grant Number SFI/12/RC/2289\textunderscore2, cofunded
by the European Regional Development Fund.

\typeout{}  
\bibliography{aaai22}

\begin{thebibliography}{18}
\providecommand{\natexlab}[1]{#1}

\bibitem[{Birhane, Prabhu, and Kahembwe(2021)}]{birhane2021multimodal}
Birhane, A.; Prabhu, V.~U.; and Kahembwe, E. 2021.
\newblock Multimodal datasets: misogyny, pornography, and malignant
  stereotypes.
\newblock \emph{arXiv preprint arXiv:2110.01963}.

\bibitem[{Buolamwini and Gebru(2018)}]{buolamwini2018gender}
Buolamwini, J.; and Gebru, T. 2018.
\newblock Gender shades: Intersectional accuracy disparities in commercial
  gender classification.
\newblock In \emph{Conference on fairness, accountability and transparency},
  77--91. PMLR.

\bibitem[{Caliskan, Bryson, and Narayanan(2017)}]{caliskan2017semantics}
Caliskan, A.; Bryson, J.~J.; and Narayanan, A. 2017.
\newblock Semantics derived automatically from language corpora contain
  human-like biases.
\newblock \emph{Science}, 356(6334): 183--186.

\bibitem[{Garg et~al.(2018)Garg, Schiebinger, Jurafsky, and Zou}]{garg2018word}
Garg, N.; Schiebinger, L.; Jurafsky, D.; and Zou, J. 2018.
\newblock Word embeddings quantify 100 years of gender and ethnic stereotypes.
\newblock \emph{Proceedings of the National Academy of Sciences}, 115(16):
  E3635--E3644.

\bibitem[{Greenwald, McGhee, and Schwartz(1998)}]{greenwald1998measuring}
Greenwald, A.~G.; McGhee, D.~E.; and Schwartz, J.~L. 1998.
\newblock Measuring individual differences in implicit cognition: the implicit
  association test.
\newblock \emph{Journal of personality and social psychology}, 74(6): 1464.

\bibitem[{https://www.facebook.com/braddwyer()}]{roboflowWhatOpenAIs}
https://www.facebook.com/braddwyer. ????
\newblock {W}hat is {O}pen{A}{I}'s {C}{L}{I}{P} and how to use it? ---
  blog.roboflow.com.
\newblock \url{https://blog.roboflow.com/openai-clip/}.
\newblock [Accessed 26-Nov-2022].

\bibitem[{Krishnakumar et~al.(2021)Krishnakumar, Prabhu, Sudhakar, and
  Hoffman}]{krishnakumar2021udis}
Krishnakumar, A.; Prabhu, V.; Sudhakar, S.; and Hoffman, J. 2021.
\newblock Udis: Unsupervised discovery of bias in deep visual recognition
  models.
\newblock In \emph{British Machine Vision Conference (BMVC)}, volume~1, 3.

\bibitem[{Mandal, Leavy, and Little(2021)}]{mandal2021dataset}
Mandal, A.; Leavy, S.; and Little, S. 2021.
\newblock Dataset diversity: measuring and mitigating geographical bias in
  image search and retrieval.

\bibitem[{Mishkin et~al.()Mishkin, Ahmad, Brundage, Krueger, and
  Sastry}]{mishkin2022risks}
Mishkin, P.; Ahmad, L.; Brundage, M.; Krueger, G.; and Sastry, G. ????
\newblock DALL·E 2 Preview - Risks and Limitations.

\bibitem[{Misra et~al.(2016)Misra, Lawrence~Zitnick, Mitchell, and
  Girshick}]{misra2016seeing}
Misra, I.; Lawrence~Zitnick, C.; Mitchell, M.; and Girshick, R. 2016.
\newblock Seeing through the human reporting bias: Visual classifiers from
  noisy human-centric labels.
\newblock In \emph{Proceedings of the IEEE conference on computer vision and
  pattern recognition}, 2930--2939.

\bibitem[{Radford et~al.(2021)Radford, Kim, Hallacy, Ramesh, Goh, Agarwal,
  Sastry, Askell, Mishkin, Clark et~al.}]{radford2021learning}
Radford, A.; Kim, J.~W.; Hallacy, C.; Ramesh, A.; Goh, G.; Agarwal, S.; Sastry,
  G.; Askell, A.; Mishkin, P.; Clark, J.; et~al. 2021.
\newblock Learning transferable visual models from natural language
  supervision.
\newblock In \emph{International Conference on Machine Learning}, 8748--8763.
  PMLR.

\bibitem[{Ramesh et~al.(2022)Ramesh, Dhariwal, Nichol, Chu, and
  Chen}]{ramesh2022hierarchical}
Ramesh, A.; Dhariwal, P.; Nichol, A.; Chu, C.; and Chen, M. 2022.
\newblock Hierarchical text-conditional image generation with clip latents.
\newblock \emph{arXiv preprint arXiv:2204.06125}.

\bibitem[{Rombach et~al.(2022)Rombach, Blattmann, Lorenz, Esser, and
  Ommer}]{rombach2022high}
Rombach, R.; Blattmann, A.; Lorenz, D.; Esser, P.; and Ommer, B. 2022.
\newblock High-resolution image synthesis with latent diffusion models.
\newblock In \emph{Proceedings of the IEEE/CVF Conference on Computer Vision
  and Pattern Recognition}, 10684--10695.

\bibitem[{Serna et~al.(2021)Serna, Pena, Morales, and
  Fierrez}]{serna2021insidebias}
Serna, I.; Pena, A.; Morales, A.; and Fierrez, J. 2021.
\newblock InsideBias: Measuring bias in deep networks and application to face
  gender biometrics.
\newblock In \emph{2020 25th International Conference on Pattern Recognition
  (ICPR)}, 3720--3727. IEEE.

\bibitem[{Sirotkin, Carballeira, and
  Escudero-Vi{\~n}olo(2022)}]{sirotkin2022study}
Sirotkin, K.; Carballeira, P.; and Escudero-Vi{\~n}olo, M. 2022.
\newblock A study on the distribution of social biases in self-supervised
  learning visual models.
\newblock In \emph{Proceedings of the IEEE/CVF Conference on Computer Vision
  and Pattern Recognition}, 10442--10451.

\bibitem[{Steed and Caliskan(2021)}]{steed2021image}
Steed, R.; and Caliskan, A. 2021.
\newblock Image representations learned with unsupervised pre-training contain
  human-like biases.
\newblock In \emph{Proceedings of the 2021 ACM conference on fairness,
  accountability, and transparency}, 701--713.

\bibitem[{Wang et~al.(2022)Wang, Liu, Zhang, Kleiman, Kim, Zhao, Shirai,
  Narayanan, and Russakovsky}]{wang2022revise}
Wang, A.; Liu, A.; Zhang, R.; Kleiman, A.; Kim, L.; Zhao, D.; Shirai, I.;
  Narayanan, A.; and Russakovsky, O. 2022.
\newblock REVISE: A tool for measuring and mitigating bias in visual datasets.
\newblock \emph{International Journal of Computer Vision}, 1--21.

\bibitem[{Wang et~al.(2019)Wang, Zhao, Yatskar, Chang, and
  Ordonez}]{wang2019balanced}
Wang, T.; Zhao, J.; Yatskar, M.; Chang, K.-W.; and Ordonez, V. 2019.
\newblock Balanced datasets are not enough: Estimating and mitigating gender
  bias in deep image representations.
\newblock In \emph{Proceedings of the IEEE/CVF International Conference on
  Computer Vision}, 5310--5319.

\end{thebibliography}

\appendix
\section{A: Image Generation Prompts}
\small

The terms and concepts used to create the text prompts are based on work by \cite{garg2018word} and \cite{wang2022revise} as a preliminary demonstrations of measuring gender bias in multimodal generative models. * denotes prompt for Stable Diffusion
\begin{table}
\small
\centering
\begin{tabular}{|p{.5\columnwidth}|p{.3\columnwidth}|}
\hline
\textbf{Prompt}                   & \textbf{Association} \\
(\emph{``an image of a/an ...''}) & \\
\hline
\multicolumn{2}{c}{~} \\  
\hline
\multicolumn{2}{|l|}{Type: Attributes, 16 images each} \\ \hline 
man & \multirow{4}{*}{male gender} \\
boy & \\
old man & \\
male young adult/teenage boy* & \\ 
\hline
woman & \multirow{4}{*}{female gender} \\
girl & \\
old woman & \\
female young adult/teenage girl* & \\
\hline
\multicolumn{2}{r}{Total attribute images: 128} \\ 

\multicolumn{2}{c}{~} \\  

\hline
\multicolumn{2}{|l|}{Target: Occupations, 20 images each} \\ \hline 
chief executive officer & \multirow{4}{*}{\shortstack[l]{male dominated \\occupations}} \\
engineer & \\
doctor & \\
farmer & \\
programmer & \\
\hline
beautician & \multirow{4}{*}{\shortstack[l]{female dominated \\occupations}} \\
housekeeper & \\
librarian & \\
secretary & \\
nurse treating a patient & \\
\hline 

\multicolumn{2}{c}{~} \\  
\hline
\multicolumn{2}{|l|}{Target: Sports, 20 images each} \\ \hline 
baseball player & \multirow{3}{*}{outdoor sports} \\
rugby player & \\
cricket player & \\ \hline
badminton player & \multirow{3}{*}{indoor sports} \\
swimmer & \\
gymnast & \\
\hline 

\multicolumn{2}{c}{~} \\  
\hline
\multicolumn{2}{|l|}{Target: Objects, 20 images each} \\ \hline 
person fixing a car & \multirow{3}{*}{male gender} \\
person operating farm machinery & \\
person with a fishing rod & \\ \hline 
person using a food processor & \multirow{3}{*}{female gender} \\
person using a hair drier & \\
person using a make-up kit & \\
\hline

\multicolumn{2}{c}{~} \\  
\hline
\multicolumn{2}{|l|}{Target: Scenes/Activities, 20 images each} \\ \hline 
person using a theodolite & \multirow{3}{*}{male gender} \\
person using a lathe machine & \\
person snowboarding & \\ \hline 
person shopping & \multirow{3}{*}{female gender} \\
person reading a romantic novel and drinking tea & \\
child playing with a dollhouse & \\
\hline
\multicolumn{2}{r}{Total target images: 560} \\ 

\end{tabular}
\end{table}

\end{document}